\documentclass[letterpaper, 10 pt, conference]{ieeeconf}  

\IEEEoverridecommandlockouts                              

\usepackage{graphicx}
\usepackage{float}
\usepackage{soul}
\graphicspath{ {./images/} } 

\usepackage{color}
\usepackage{amsmath}
\DeclareMathOperator*{\argmax}{argmax}
\usepackage{breqn}

\usepackage[usenames,dvipsnames]{xcolor}

\usepackage{algorithm}
\usepackage[noend]{algpseudocode}
\algnewcommand{\LineComment}[1]{\State\textcolor{RawSienna}{\(\triangleright\) #1}}

\usepackage{balance}
\usepackage{amsfonts}
\usepackage{amssymb}
\usepackage{array}
\newcolumntype{P}[1]{>{\centering\arraybackslash}p{#1}}

\usepackage{tikz}
\usetikzlibrary{matrix, positioning}

\usepackage{booktabs} 

\title{\LARGE \bf
Stochastic Assignment for Deploying Multiple Marsupial Robots 
}
\author{Chris (Yu Hsuan) Lee, Graeme Best, Geoffrey A. Hollinger
\thanks{*Approved for public release; distribution is unlimited. This work was in part sponsored by DARPA under agreement \#HR00111820044 and Office of Naval Research Grant \mbox{N00014-17-1-2581}. Any opinions, findings and conclusions or recommendations expressed in this material are those of the authors and do not necessarily reflect those of the sponsor.}
\thanks{*The authors are with the Collaborative Robotics and Intelligent Systems (CoRIS) Institute, Oregon State University, Corvallis OR, USA. 
	{\tt\small \{leeyuh,bestg,geoff.hollinger\}@oregonstate.edu}}%
}

\begin{document}

\maketitle
\thispagestyle{empty}
\pagestyle{empty}

\begin{abstract}
Marsupial robot teams consist of \textit{carrier} robots that transport and deploy multiple \textit{passenger} robots, such as a team of ground robots that carry and deploy multiple aerial robots, to rapidly explore complex environments. We specifically address the problem of planning the deployment times and locations of the carrier robots to best meet the objectives of a mission while reasoning over uncertain future observations and rewards. While prior work proposed optimal, polynomial-time solutions to \textit{single}-carrier robot systems, the \textit{multiple}-carrier robot deployment problem is fundamentally harder as it requires addressing conflicts and dependencies between deployments of multiple passenger robots. We propose a centralized heuristic search algorithm for the multiple-carrier robot deployment problem that combines Monte Carlo Tree Search with a dynamic programming-based solution to the Sequential Stochastic Assignment Problem as a rollout action-selection policy. Our results with both procedurally-generated data and data drawn from the DARPA Subterranean Challenge Urban Circuit show the viability of our approach and substantial exploration performance improvements over alternative algorithms.
\end{abstract}

\section{Introduction}



Rapid robotic exploration of a wide-range of environments, such as for urban search and rescue~\cite{moore2016nested,stankiewicz2018motion,zhao2008research} or marine monitoring~\cite{das2015data,kalaitzakis2020marsupial,mccammon2021ocean}, can be achieved by heterogeneous multi-robot systems called \textit{marsupial robots}. These systems typically consist of a large and robust \textit{carrier} robot (e.g., a ground vehicle or surface vessel) that transports smaller but agile \textit{passenger} robot(s) (e.g., aerial or underwater robots)~\cite{Murphy1999}. By utilizing complementary capabilities between the different types of robots, marsupial robot systems are able to explore complex environments that contain navigation or sensing challenges, such as vertical shafts or smoky occlusions, that a single type of robot cannot overcome on its own. As we scale to larger and more complex environments, it is necessary to expand these marsupial robot teams to comprise multiple carrier robots. This enables carrying larger teams of passenger robots and achieving a wider spatial coverage of passenger robot deployment locations.


A key challenge to address in these systems is the selection of a high value set of passenger robot deployment locations. The value of deployment locations are quantified through application-specific rewards, such as frontier cells for exploration tasks. This problem requires reasoning over the value of deploying a passenger robot immediately at the current decision point compared to the uncertain value of possible future deployment locations~\cite{lee2021optimalSSAP}. Additionally, as we scale to larger numbers of passenger robots, it is also necessary to ensure the \textit{set} of selected deployment locations do not include conflicting deployments within close proximity to each other that would prohibit efficient and safe exploration.

For the case of a \textit{single} carrier robot, an optimal, polynomial-time deployment planning algorithm was proposed in our previous work~\cite{lee2021optimalSSAP}. This algorithm draws upon the close relationship to the Sequential Stochastic Assignment Problem (SSAP)~\cite{derman1972sequential}, which poses the problem of selecting an optimal subset of a set of sequentially-revealed random variables. The decision to include or exclude a random variable must be made as it is revealed while reasoning over prior knowledge of future rewards. This SSAP decision process is analogous to the process of deploying marsupial robots.

However, this SSAP-based algorithm~\cite{lee2021optimalSSAP} fails to adequately address the complexities of the \textit{multiple} carrier robot deployment problem. Specifically, it is suboptimal in scenarios where it is necessary to reason over conflicts between possible deployment locations, which arise due to, for example, the need to not deploy passenger robots in close proximity. We address this generalized problem in this paper.


We propose the first solution algorithm to the multiple-carrier, multiple-passenger marsupial robot deployment problem. At each candidate deployment location, the algorithm reasons over the value of deploying now in comparison to the expected future value of continuing on and deploying at a later stage. This reasoning is performed by employing Monte Carlo Tree Search (MCTS)~\cite{browne2012survey} coupled with a fast and effective problem-specific heuristic rollout policy. MCTS optimizes over the joint deployment action space of the multiple carrier robots while considering penalties introduced by deployment conflicts. Rather than using the standard, naive random rollout policy within MCTS, we propose an action selection heuristic that generalizes the single carrier robot SSAP algorithm~\cite{lee2021optimalSSAP}. This rollout policy influences the subsequent tree search expansions, and thus improves the overall performance of the tree search. The algorithm is anytime and maintains the asymptotic convergence guarantees of MCTS~\cite{kocsis2006uctB}. Our MCTS algorithm is executed by a centralized decision maker, motivated by scenarios where sufficient communication infrastructure is available; however, we also anticipate the methods proposed here to be a necessary foundation for the development of decentralized planners for this problem in the future.


We present simulated marsupial robot deployment experiments where a team of carrier ground robots housing multiple aerial passenger robots seeks to deploy the passenger robots in deployment locations with high exploration value. Each carrier robot observes features of interest that are unreachable to the ground robots but accessible to passenger aerial robots. We present results in two scenarios: one with random features of interest drawn from a Poisson point process, and the other where these features are derived from a dataset recorded by three exploring robots during the DARPA Subterranean Challenge Urban Circuit~\cite{scherer2021fr}. In experiments with up to 6 carrier robots and 18 passenger robots, our algorithm outperforms the optimal deployment policy to the relaxed deployment problem without conflicts~\cite{lee2021optimalSSAP} and MCTS with the standard random rollout policy.



\section{Related Work}

The increased complexity of marsupial robots over more traditional multi-robot systems require planners that are able to handle tasks such as coordination, deployment, and retrieval of the passenger robots, in addition to typical planning requirements of multi-robot systems. Planners must consider locations that are traversable by the carrier robot and obstacles that are not, requiring the differentiating capabilities of passenger robots~\cite{wurm2013coordinating,schwager2011unifying}.

The problem of planning deployment locations is key to the success of all of the above decision making, as a poor deployment location selection can result in passenger robots being unable to contribute towards the mission objectives. These deployment decisions have largely been left to human operators~\cite{schuster2020arches}, who may not have adequate situational awareness to effectively make these decisions. One way to automate these decisions is to formulate a policy that triggers deployments based on predefined Boolean conditions, such as the discovery of a goal that is unreachable to the carrier ground robot~\cite{stankiewicz2018motion,las2015path}. However, these relatively simple approaches are limited in applicability and do not reason over the long term value of decisions, particularly when there are multiple passenger robots to deploy.

These limitations have been addressed by a recently proposed SSAP-based algorithm~\cite{lee2021optimalSSAP}. The SSAP algorithm reasons over the observations encountered in the environment and possible future observations in order to deploy multiple passenger robots in high value locations for robotic information gathering. Other related deployment algorithms include formulating the problem as a coverage-type problem~\cite{mei2006deployment,hansen2018autonomous}, as well as algorithms that combine more complex triggers such as power consumption and communication availability~\cite{lee2018path}. However, these approaches do not consider conflicts and dependencies between different deployment decisions. We leverage MCTS to reason over these conflicts and dependencies while borrowing ideas from the SSAP-based algorithm~\cite{lee2021optimalSSAP}.

MCTS~\cite{browne2012survey} is rising in popularity as a high performing search algorithm for sequential decision making in robotics. This is in part due to its flexibility to optimize with respect to a general class of reward functions and its ability to naturally handle uncertainty~\cite{mccammon2021ocean, chen2019pareto, best2019decmcts, bauer2019monte, corah2019distributed, kartal2015stochastic}. Best results are often achieved when replacing the default random rollout policy with problem-specific heuristics, which are directly leveraged to guide the best-first search procedure~\cite{james2017analysis}. In robotics, a common problem-specific rollout policy is the greedy policy for information gathering~\cite{mccammon2021ocean, best2019decmcts, lauri2015mcts}. For the deployment problem, the single-carrier SSAP solution discussed above~\cite{lee2021optimalSSAP} presents a useful heuristic for the multi-carrier generalization.

MCTS-based algorithms have also recently been proposed for the context of multi-robot systems. These include centralized algorithms where a single search tree represents the action space of all robots~\cite{kartal2015stochastic}, and decentralized algorithms where each robot separately searches over a tree representing its own actions~\cite{best2019decmcts, corah2019distributed}. Here, we focus on centralized contexts, and thus formulate a similar search tree to~\cite{kartal2015stochastic}, but our proposed SSAP-based rollout policy is also applicable to the decentralized MCTS variants.

\section{The Multiple Marsupial Robot\\Deployment Problem}
\label{sec:problem_formulation}

\begin{figure}[tp]
    \centering
    \includegraphics[width=\linewidth]{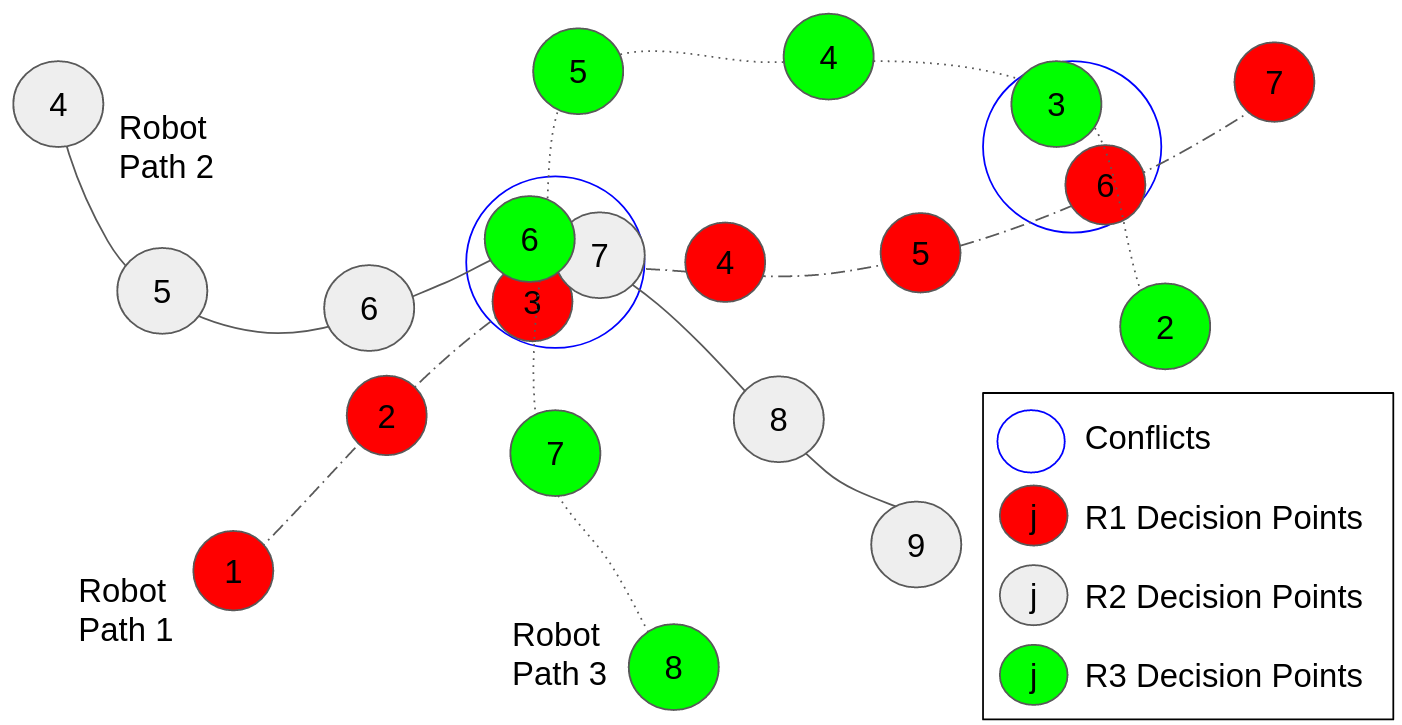}
    \caption{A deployment problem example with $R=3$ carrier robots. Each colored circle represents a deployment decision location with the associated stage $j$. Here, overlap conflicts can be seen inside the blue circles. The first overlap conflict would be defined as $o_1 = \{[r_1,3],[r_2,7],[r_3,6]\}$ and the second conflict defined as $o_2 = \{[r_1,6],[r_3,3]\}$.}
    \label{fig:problem_formulation}
\end{figure}
We consider a marsupial robot system consisting of multiple carrier robots, each carrying multiple passenger robots. A centralized decision maker must make online decisions regarding when to deploy the passenger robots. These decisions must be made while reasoning over both the reward gained from deploying immediately and the estimated rewards for deploying at an unknown later deployment location. The decision maker must also consider potential deployment conflicts by reasoning over the loss of reward caused by multiple deployments in close proximity. We formalize this problem as follows.

\subsection{Marsupial Robot Team Decision Points}
\label{sec:decision_points}

The multi-robot team consists of $R$ carrier robots denoted $(r_1, r_2, ..., r_R)$, with each carrier robot carrying $D$ passenger robots to deploy. Along each carrier robot's path, there are $N$ stages that represent possible deployment locations. For each carrier robot $r$, there are a sequence of $N$ random variables $(X_1^r, X_2^r, ..., X_N^r)$, which represent deployment rewards associated with each decision point (ignoring the conflicts introduced below in Sec.~\ref{sec:conflicts}). We assume the random variables $x$ are distributed according to a known prior distribution $f(x)$ and are assumed to be independent. At stage $j \in \{ 1,...,N \}$, the carrier robot reaches a decision point, and the outcomes of all random variables $(X_1^r, X_2^r, ..., X_j^r)$, denoted $(x_1^r, x_2^r, ..., x_j^r)$, have been observed and are known to the decision maker. 

For each carrier robot, a subset of the $N$ stages may be invalid deployment locations; for example, due to staggered start times between the carrier robots. To maintain consistent indexing between the different carrier robots, we denote the realized observations at an invalid deployment stage $j$ for a robot $r$ as $x_j^r=\varnothing$.


\subsection{Deployment Conflicts}
\label{sec:conflicts}

As illustrated in Fig.~\ref{fig:problem_formulation}, some possible deployment locations may be in close proximity to each other. However, in many applications, it is undesirable to deploy passenger robots in close proximity to each other since it may, for example, lead to inefficient exploration of the environment. We refer to groups of decision points in close proximity with each other as \textit{conflicts}. Each conflict is defined by an overlap $o \in \mathcal{O}$, where $\mathcal{O}$ is the set of all overlaps. Each overlap $o \in \mathcal{O}$ is defined as a set of robot--stage pairs $o = \{ [r_1,j_1], [r_2,j_2],...\}$, which implies that the deployment location for carrier robot $r_1$ associated with stage $j_1$ is in conflict with the robot--stage pairs with $o$. For simplicity, we focus on addressing problems with overlap definitions where each carrier robot $r$ does not have conflicts with itself at later stages (i.e., $r_i \ne r_j, \forall [r_i, j_i] \ne [r_j, j_j] \in o$), although the proposed algorithm allows for relaxing this constraint.

If carrier robot $r$ deploys a passenger robot at stage $j$, then a reward of $v_d^r = x_j^r/p$ is received, where $1/p$ is a penalty factor associated with conflicts. Specifically, we define $p$ as the number of selected passenger robot deployments that are within the same conflict set as $[r,j]$. This penalty factor defines a loss of reward associated with making multiple deployments in close proximity (conflict) with each other. We note that while we focus on this conflict penalty definition throughout this paper, our proposed algorithm would also be applicable to other definitions. 


\subsection{Deployment Actions}
The action of carrier robot $r$ at stage $j$ is defined as $a_j^r$. At each deployment location, carrier robot $r$ has two possible actions: deploy the passenger robot, or continue to the next deployment location. The joint actions of all of the $R$ carrier robots at stage $j$ is defined as $A_j = \{a_j^1, a_j^2,...,a_j^R\}$. If the carrier robot has no remaining passenger robots to deploy, the carrier robot defaults to a no-deploy action. At stage $j$, the decision maker knows the previous deploy actions of all carrier robots.


\subsection{Problem Statement}

The set of stage indices where the passenger robots are deployed by carrier robot $r$ are defined as $d^r = (d_1^r, d_2^r,...,d_D^r)$. Furthermore, the set of all of the deployed stage indices across all carrier robots is defined as $\pi = \{d^1, d^2,...,d^R\}$. As such, the goal of the decision maker is to maximize the expected sum of the rewards returned from the chosen deployment locations; i.e., find the optimal deploy sequences
\begin{align}
\label{eqn:global_objective}
    \pi^* = \argmax_{\pi}\mathbb{E} \Big[\sum_{r=1}^R\sum_{d=1}^D v_d^r \Big],
\end{align}
where $R$ is the number of carrier robots, $D$ is the number of passenger robots per carrier robot, and $v_d^r$ is the reward for each deployment after accounting for conflicts (penalty factor $1/p$). The expectation in~\eqref{eqn:global_objective} is with respect to the unknown outcomes $(x_1^r, x_2^r, ..., x_j^r)$ of random variables $(X_1^r, X_2^r, ..., X_j^r)$. 

The key computational challenges presented by this problem lie in the need to reason over these unknown rewards of future deployment decisions and also the need to account for deployment conflicts, which lead to dependencies in the reward structure.

\section{MCTS with Sequential Stochastic Assignment Rollout Policy}
\label{sec:approach}

We propose a centralized algorithm for the multiple marsupial robot deployment problem that combines MCTS~\cite{browne2012survey} with heuristics derived from solutions to SSAP~\cite{derman1972sequential}. MCTS is a general-purpose sequential decision making algorithm that admits problem-specific heuristics. While SSAP-based algorithm have been proposed as an optimal solution to the \textit{single} carrier robot deployment problem~\cite{lee2021optimalSSAP}, here a SSAP-based algorithm serves as a strong heuristic to guide our MCTS-based algorithm for the \textit{multiple} carrier robot deployment problem, which requires additionally addressing deployment conflicts. 

This section begins by providing an overview of the tree data structure and overall algorithm, followed by a detailed description of the key algorithmic components and a brief analysis of runtime complexity and optimality.




\subsection{Search Tree and Algorithm Overview}
\label{sec:methods_alg_overview}

At stage $j$, our online algorithm incrementally expands and searches over a search tree $\mathcal{T}_j$, which represents the search space for the remaining deployment decisions. The root of $\mathcal{T}_j$ represents stage $j$, while subsequent layers of $\mathcal{T}_j$ are associated with future stages $\{j+1,...,N\}$. The outgoing edges of a vertex enumerate all possible joint actions $A_i$ available to the team at stage $i$.

Pseudocode for the algorithm is provided in Alg.~\ref{alg:pseudocode_1}. The search proceeds by employing standard MCTS~\cite{browne2012survey} (described further in Sec.~\ref{sec:methods_mcts}) with an adapted rollout phase that exploits information provided by the optimal solution to the single carrier deployment problem to guide the search tree expansion (described further in Sec.~\ref{sec:methods_ssap_rollout}). 
It does this while reasoning over the various inputs introduced in Sec.~\ref{sec:problem_formulation}, which we summarize as follows:
\begin{itemize}
    \item the current observations $x_j^r$ by each carrier robot of the associated random variables,
    \item the prior belief distribution $f(x)$ for all possible future deployment locations,
    \item the previous actions of the carrier robots $(A_1, ...,A_{j-1})$, which influences potential future deployment conflicts,
    \item the set of deployment conflict overlaps $\mathcal{O}$, and
    \item the number of stages remaining: $N-j$.
\end{itemize}
The search concludes after a predefined number of iterations and returns the estimated best joint deployment action $A_j$ to execute at stage $j$.



\begin{algorithm*}[t]
\caption{Overview of the online deployment algorithm for selecting the deployment actions $A_j$ at stage $j$.}
\label{alg:pseudocode_1}

\begin{algorithmic}[1]
\State $\mathcal{T}_j \gets $ initialize MCTS tree
\LineComment{Incrementally expand the search tree $\mathcal{T}_j$}
\For{fixed number of iterations}
    \State $n \gets $ \Call{SelectNode}{$\mathcal{T}_j$} \Comment{Select node to expand using UCT~\cite{kocsis2006uctB}}
    \State $n^+ \gets $ \Call{ExpandTree}{$n$} \Comment{Add new child node}  
    \State $(x_{j+1}^r, ..., x_N^r)_{\forall r} \gets $ \Call{SampleFutureObservations}{$f$} \Comment{Sample from the prior belief distribution}
    \State $(A_j, ..., A_N) \gets $ \Call{SSAP\_Rollout}{$n^+, (A_1, ..., A_{j-1}), (x_{j+1}^r, ..., x_N^r)_{\forall r}, \mathcal{O}$} \Comment{SSAP heuristic: See Sec.~\ref{sec:methods_ssap_rollout}}    
    \State $v \gets $ \Call{EvaluateSolution}{$(x_1^r, ..., x_N^r)_{\forall r}, (A_1, ..., A_N), \mathcal{O}$} \Comment{Evaluate the selected action sequence}
    \State \Call{BackpropagateReward}{$n^+, v$} \Comment{Update statistics along the path back to the root node}
\EndFor
\LineComment{Extract the solution}
\State $n^* \gets $ best node in $\mathcal{T}_j$
\State $A_j \gets $ first joint action in sequence for $n^*$
\State \Return $A_j$
%
    
        
%
\end{algorithmic}
\end{algorithm*}

\subsection{MCTS for Multiple Carrier Robot Deployments}
\label{sec:methods_mcts}

Our deployment algorithm employs the standard MCTS algorithm~\cite{browne2012survey} to incrementally expand and search over the search tree $\mathcal{T}_j$. We summarize MCTS as follows, and defer our main innovation---the inclusion of a new SSAP-based rollout policy---to Sec.~\ref{sec:methods_ssap_rollout}. MCTS consists of four phases: \textit{selection}, \textit{expansion}, \textit{rollout}, and \textit{backpropagation}. 

In the \textit{selection} phase (Alg.~\ref{alg:pseudocode_1} line~4) a path is followed through $\mathcal{T}_j$ from the root node to a leaf node $n$. As per the standard UCT selection policy~\cite{kocsis2006uctB}, this path is created by recursively selecting child nodes that maximize an upper confidence bound:
\begin{equation}\label{eqn:ucb}
\bar{v}(n^\prime) + c \sqrt{\frac{\ln t(n) }{t(n^\prime)}},
\end{equation}
where $\bar{v}(n^\prime)$ is an estimate of the average reward associated with the node $n^\prime$ that is being considered, $t(n)$ is the number of samples that have passed through the parent node $n$, $t(n^\prime)$ is the number of samples that have passed through the considered node $n^\prime$, and $c$ is the exploration parameter (typically set as $\sqrt{2}$ if the rewards are scaled between 0 and 1). The purpose of following this upper confidence bound policy is to naturally balance between exploring the tree (i.e., pick nodes that have been considered fewer times) and exploiting the knowledge gained so far (i.e., pick nodes with high expected rewards).

In the \textit{expansion} phase (line~5), a new child node $n^+$ is added to $\mathcal{T}_j$ that extends from the selected node $n$. This node $n^+$ is chosen arbitrarily from any of the feasible joint actions that may be performed from the state represented by $n$. For our context of the deployment problem, a joint action is considered feasible if all carrier robots do not deploy more than the number of passenger robots they are carrying, deployments only occur at valid deployment stages (see Sec.~\ref{sec:decision_points}), and at stage $N$ all passenger robots have been deployed.

In the \textit{rollout} phase (lines~6--8), a sequence of feasible actions is generated according to some given policy and then evaluated with respect to the reward function. In standard MCTS, the given policy is a random policy. However, there is also an opportunity here to employ problem-specific heuristics here to improve the reward estimates and speed up the search. We discuss our proposed rollout policy below in Sec.~\ref{sec:methods_ssap_rollout}.

Every node $n$ in the tree maintains a reward average $\bar{v}(n)$ and a count $t(n)$ of the number of rollouts that pass through this node. These two quantities are updated during the \textit{backpropagation} phase (line~9). The rollout evaluation $v$ is merged into the averages stored at all nodes on the path from the expanded node $n^+$ back to the root node, and the associated counts are incremented. These updated statistics guide the selection phase in the following iteration, and thus the tree expansions are directly influenced by the rollout evaluations.

At the end of the search (lines 10--13), the best node $n^*$ is picked as the node with the highest expected reward. The first joint action in the sequence associated with $n^*$ defines the deployment actions to be executed at this stage $j$. The algorithm is online such that the algorithm is run again at stage $j+1$ while using any new gathered information, particularly the previous deployment decisions and new observations.

\subsection{SSAP Action Selection Rollout Policy}
\label{sec:methods_ssap_rollout}

As discussed above, problem-specific heuristics can be used within the \textit{rollout} phase to improve the reward estimates, which in turn improves the MCTS node selection and therefore the overall search performance. Ideal rollout policies should be both fast to compute and provide good reward estimates. For our deployment problem, we draw inspiration from a solution to the single carrier robot problem presented in~\cite{lee2021optimalSSAP}. While it is an optimal solution to the single carrier formulation, for the multi carrier problem this strong guarantee does not hold due to the need to address the conflicts, which breaks the independence assumption. However, we adapt it to provide reasonable reward estimates for our generalized problem that can be computed quickly within the online search algorithm.

Both the single carrier solution~\cite{lee2021optimalSSAP} and our adapted heuristic for the multiple carrier case are based on the solution to the Sequential Stochastic Assignment Problem (SSAP)~\cite{derman1972sequential}. The aim of SSAP is to find the optimal subset (with given cardinality) of a set of random variables whose realizations are revealed sequentially. The decision to include or exclude a random variable from the subset must be made immediately when its realization is revealed.  

SSAP maps closely to our deployment problem, as the value of a deployment location is only revealed as the carrier robot arrives at a location, and decisions to deploy or not deploy at a decision point are irreversible. We briefly formalize the optimal solution to SSAP as follows (further details may be found in~\cite{lee2021optimalSSAP, derman1972sequential}) and discuss how we adapt it to be used as an MCTS rollout policy.

The optimal SSAP policy~\cite{derman1972sequential} consists of precomputing a set of thresholds $\{a_{i,n}\}$ for each stage. The realization $x_j^r$ of the random variable $X_j^r$ is compared to the relevant thresholds to determine whether to deploy now or wait until a later decision point. These thresholds are computed by considering the number of stages $N$ and the prior belief distribution for the rewards $f(x)$. Specifically, for $n=N-j+1$ stages remaining, the thresholds are computed via the recurrence relationship
\begin{align}
\label{eqn:ssap_threshold_recurrence}
    a_{i,n+1} &= a_{i-1,n} \int_{-\infty}^{a_{i-1,n}}f(x)dx \nonumber
    +\int_{a_{i-1,n}}^{a_{i,n}}xf(x)dx \nonumber \\
    &\quad \quad \quad 
    +a_{i,n} \int_{a_{i,n}}^{\infty}f(x)dx,
\end{align}
where $-\infty\cdot0 = 0$ and $\infty\cdot0 = 0$. The second term represents an expectation of reward for the case where the reward is revealed to be within the two thresholds and therefore the decision maker receives this reward. The first integral represents the expected future reward if the reward lies in a lower threshold, and the third integral is for the case where it lies in a larger threshold. The set of all thresholds can be computed in quadratic time. If there are $n-i$ passenger robots left to deploy at stage $j=N-n+1$, then the optimal policy is to deploy if $x_j^r$ is greater than the $i$th interval, $a_{i,n}$, and do not deploy otherwise.
%

However, while the SSAP deployment policy is optimal for a single carrier robot, it does not consider multiple carrier robots and potential conflicts between the deploy locations, as defined in Sec.~\ref{sec:conflicts}. The conflicts can be partly accounted for by adjusting the belief reward distributions $f(x)$ at decision points that conflict with \textit{known} prior deployments by multiplying by the $1/p$ penalty factor. However, addressing unknown future conflicts would require enumerating all permutations of future deployments, which is infeasible to achieve within a fast MCTS rollout policy. Therefore, we instead rely on future MCTS iterations to correct for any error in the reward estimates by this SSAP heuristic.

The MCTS rollout policy (Alg.~\ref{alg:pseudocode_1} lines 6--8) also requires evaluating the reward that will be achieved by the selected sequence of actions. While for specific belief distribution structures, it may be possible to efficiently compute the expectations in \eqref{eqn:global_objective} for a specific deployment sequence, in general this will be impractical. Instead, we suggest approximating this expectation via Monte Carlo sampling of the random variable realizations from the prior distribution $f(x)$. The actions executed by a rollout can be selected by following the SSAP policy with respect to a sampled realization of the random variables $(X_{j+1}^r, ..., X_N^r), \forall r$. Within MCTS, these reward estimates are then used within the backpropagation phase (line~9) to guide the subsequent tree expansions.

\subsection{Analysis}
\label{sec:analysis}
The fully expanded search tree contains $2^{RN}$ tree nodes, where $R$ is the number of robots, and $N$ is the number of stages. MCTS is an iterative anytime algorithm that expands a subset of this search space. Thus, the computation time is directly proportional to the number of iterations, which in practice is typically selected by the user. The runtime complexity of each MCTS iteration of our algorithm is dominated by the SSAP rollout computation, which is achieved in quadratic time in the number of stages remaining~\cite{lee2021optimalSSAP}, and the reward evaluation, which has runtime proportional to the number of samples used in the Monte Carlo sampling. 

MCTS also provides strong convergence rate guarantees for the convergence to the optimal action selection~\cite{kocsis2006uctB}. In practice, MCTS performance can be further enhanced with the use of problem-specific rollout policy heuristics~\cite{james2017analysis}. While the proposed SSAP policy is optimal in the single-carrier case with no conflicts, this optimality guarantee does not extend to the generalized multiple-carrier case we consider here as the conflicts break the independence assumption of the reward distributions. Therefore, SSAP is instead best used as a strong heuristic within a general search algorithm that provides optimality guarantees, such as MCTS as we have proposed here.

We note that the search tree $\mathcal{T}_j$ that is incrementally expanded by our algorithm branches on actions but not observations. Thus, the convergence guarantee is toward the optimal open-loop action sequence rather than closed-loop policy. The advantage of seeking the open-loop action sequence is a lower branching factor, although it may come at the cost of optimality in some contexts. Alternative MCTS algorithms have been proposed for addressing the policy search problem~\cite{silver2011pomcp}, and our SSAP rollout policy could readily be used within these generalized MCTS algorithms. 








\section{Experiments and Results}
\label{sec:experiments}
We evaluate our multi-carrier, multi-passenger robot deployment algorithm in deployment experiments with Poisson-distributed simulated data and real-world data from the DARPA Subterranean Challenge~\cite{allen_chung}. The goal in these scenarios is for the passenger robots to maximally observe a set of features of interest by deploying passenger robots at high value locations.
Our deployment algorithm is compared to other deployment strategies in order to demonstrate the benefits of accounting for the conflicts and employing a problem-specific MCTS rollout policy.

\subsection{Comparison Methods}

In the following experiments, we compare the following deployment strategies:
\begin{itemize}
    \item \textit{MCTS SSAP Rollout}: Our method as described in Sec.~\ref{sec:methods_alg_overview} with the rollout policy described in Sec.~\ref{sec:methods_ssap_rollout}.
    \item \textit{MCTS Random Rollout}: A variant of our method, however with the standard random policy~\cite{browne2012survey} during the \textit{rollout} phase.
    \item \textit{Single Robot SSAP}: An optimal online deployment strategy for a single carrier robot with multiple passenger robots, adapted from~\cite{lee2021optimalSSAP}. This deployment strategy is applied to each carrier robot independently, without consideration for the conflicts between the deployment locations.
    \item \textit{Random}: A baseline method that selects $D$ decision points randomly from the path of each carrier robot. 
\end{itemize}

\subsection{Capturing Poisson-Distributed Features of Interest}
\label{sec:simulated_dat}
\subsubsection{Experimental Setup}
A system of multiple marsupial robots, composed of carrier robots transporting multiple passenger robots, travels through a 2D simulated world. As each carrier robot travels along the path, the robot detects features of interests at each decision point $j$. The features of interest in the environment are distributed as a stationary Poisson point process. Accordingly, the number of features $x_j^r$ detected by carrier robot $r$ at stage $j$ is modeled as a Poisson distribution with probability mass function
\begin{equation}
    \label{eqn:poisson_distribution}
    f(x) = \frac{\lambda^{x}e^{-\lambda}}{x!}, \text{ for } x \in \{0,1,...\}.
\end{equation}
If the carrier robot decides to deploy at the decision point, a reward of $v_j^r = x_j^r/p$ is received. 
The set of overlaps $\mathcal{O}$ were randomly generated and kept constant throughout the trials. 

To ensure the rewards were approximately in the range $[0,1]$, the rewards were normalized by the 90th percentile of the belief distribution $f(x)$. The exploration constant $c$ was set as $0.05 \times \sqrt{2}$, which encouraged more exploitation of the problem-specific rollout function.

\subsubsection{Results}
\begin{figure}[tp]
    \centering
    \includegraphics[width=\linewidth, trim=6mm 0mm 12mm 6mm, clip]{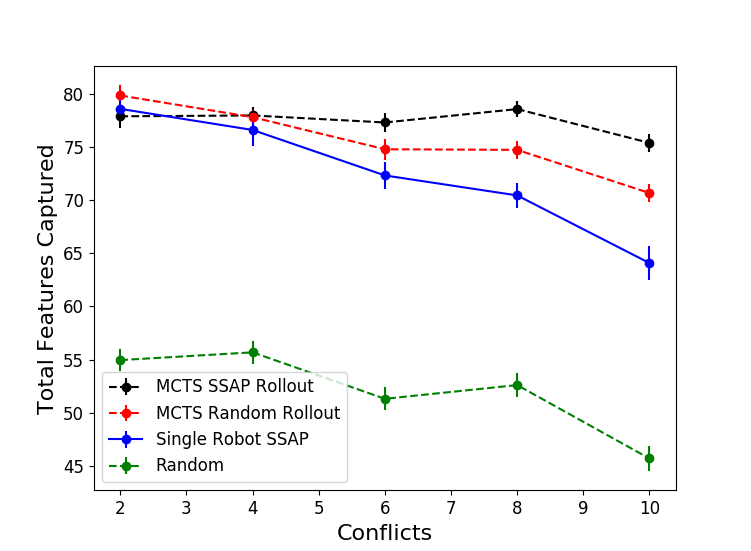}
    \caption{Algorithm comparisons with error bars (one standard error of the mean) of $R=4$ carrier robots, each deploying $D=3$ passenger robots, and traveling for $N=10$ stages over 50 trials. At each stage, MCTS ran for 10,000 iterations. The performance of \textit{Single Robot SSAP} suffers as conflicts increases, due to its inability to account for overlaps. \textit{MCTS SSAP Rollout} algorithm leverages the SSAP heuristic to improve over \textit{MCTS Random Rollout} as the number of overlap conflicts increases, making more efficient use of the limited number of iterations. }
    \vspace{-1ex}
    \label{fig:sim_performance_comparison}
\end{figure}

\begin{figure}[tp]
    \centering
    \includegraphics[width=\linewidth, trim=6mm 0mm 12mm 6mm, clip]{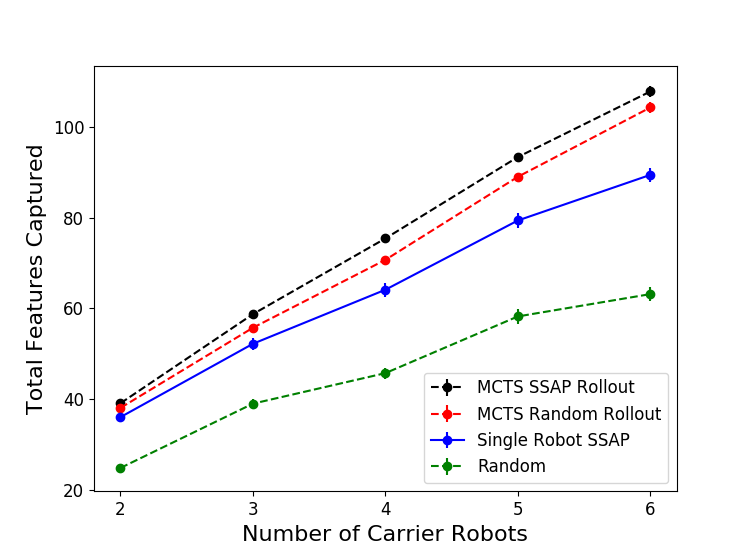}
    \caption{Algorithm comparisons with error bars (one standard error of the mean) of various numbers of carrier robots, each deploying $D=3$ passenger robots. The robots travel for $N=10$ stages and encounter 10 potential overlap conflicts over 50 trials. At each stage, MCTS ran for 10,000 iterations.}
    \vspace{-1ex}
    \label{fig:scaling_comparison}
\end{figure}


Fig.~\ref{fig:sim_performance_comparison} shows results for simulated trials with $R=4$ carrier robots, each carrying $D=3$ passenger robots. 
Both the \textit{MCTS SSAP Rollout} and \textit{MCTS Random Rollout} algorithm captured more features than the \textit{Single Robot SSAP} algorithm. The performance difference showcases the advantages of MCTS in accounting for overlap conflicts between the carrier robots and avoiding reward penalties stemming from deployment conflicts that the \textit{Single Robot SSAP} is unable to do. 
As the number of conflicts increases, \textit{MCTS SSAP Rollout} outperforms the \textit{MCTS Random Rollout} since the SSAP rollout heuristic is better able to reason over potential passenger deployment conflicts. All methods consistently outperformed \textit{Random}.

We also show the scaling of our algorithm as the number of carrier robots $R$ increases from $R=2$ to $R=6$. The number of MCTS iterations was kept constant across all trials. Despite the increasingly large joint action space, the MCTS algorithms consistently outperform the competing algorithms. By leveraging the search capabilities of MCTS, both \textit{MCTS SSAP Rollout} and \textit{MCTS Random Rollout} are able to deploy large amounts of passenger robots without heavy penalties incurred from conflicts. 
The potential for conflicts and overlap penalties increases with more robots in the environment, hindering the performance of \textit{Single Robot SSAP}, which only considers each robot independently and does not account for overlap conflicts. 

\subsection{Subterranean Challenge Urban Circuit Data}
To provide a real-world scenario for the application of the algorithm, we tested the efficacy of our deployment algorithm on recorded data from the Urban Circuit of the DARPA Subterranean Challenge~\cite{allen_chung}. A team of marsupial robots consisting of ground robots carrying aerial robots autonomously explore the Satsop Nuclear Power Plant in Elma, Washington. At the competition, several teams utilized marsupial robots, but the aerial robot deployment decisions were manually initiated by the human operator~\cite{scherer2021fr}. The operator needed to constantly look for valuable deployment locations for the passenger drones to deploy. As such, the operator was required to constantly split their attention between other tasks and identifying ideal deployment locations. This placed an additional burden on the operator, on top of many other operational tasks required in the competition. Our algorithm aims to directly address the deployment problem autonomously, in order to relieve the burden from the operator. 

\subsubsection{Experimental Setup}
Frontier cells in an occupancy grid were selected as a representation of features of interest (e.g., shafts, ledges, passageways) that are worthy for the passenger aerial robot to explore but cost-prohibitive to access by the ground robot. Using OpenVDB~\cite{museth2013openvdb}, an occupancy grid of the world was generated from the LiDAR data collected from the competition. The value of a deployment location is based on the total number of frontier cells captured within a 10\,m radius of a ground robot's location. Cells are considered to be frontier cells if they are a free cell neighboring at least one unknown cell~\cite{yamauchi1997frontier}. Additionally, we filter out easily accessible frontier cells by the ground robot and only retain frontier cells which are above 1\,m and below 0.1\,m the ground robot's position.

The carrier robots were deployed from the start location asynchronously and deployment decision locations were set every 100 seconds from the start of the hour-long run. The reward sequences for each robot were padded with $\varnothing$ observations for times where the carrier robot had not yet deployed from the start location, terminated the exploration early, or the robot's progress was stalled. For example, at stage $j=3$, time-from-start $t=300$, robot R1 has already started to explore the environment but robot R2 is still at the start gate, awaiting to be deployed into the environment. The observations of robot R1 at stage $j=3$ would be $\{x_1^1, x_2^1, x_3^1\}$, whereas for robot R2, it would be $\{\varnothing, \varnothing, \varnothing\}$. For stages where robot $r$ has an $\varnothing$ observation value, the deployment algorithm removes the child nodes containing deployment actions by robot $r$ in the \textit{expansion} phase and modifies the SSAP threshold calculations in the \textit{rollout} phase to only consider $N'$ stages remaining, where $N'$ is the number of total non-$\varnothing$ stages. For the overlap conflicts between all three robots, any deployment points within 30\,m (3 times the 10\,m observation radius) of each other were considered to be conflicting and added to the overlap set $\mathcal{O}$.

\subsubsection{Results}
\begin{figure}[tp]
    \centering
    \includegraphics[width=\linewidth, trim=0mm 0mm 0mm 0mm, clip]{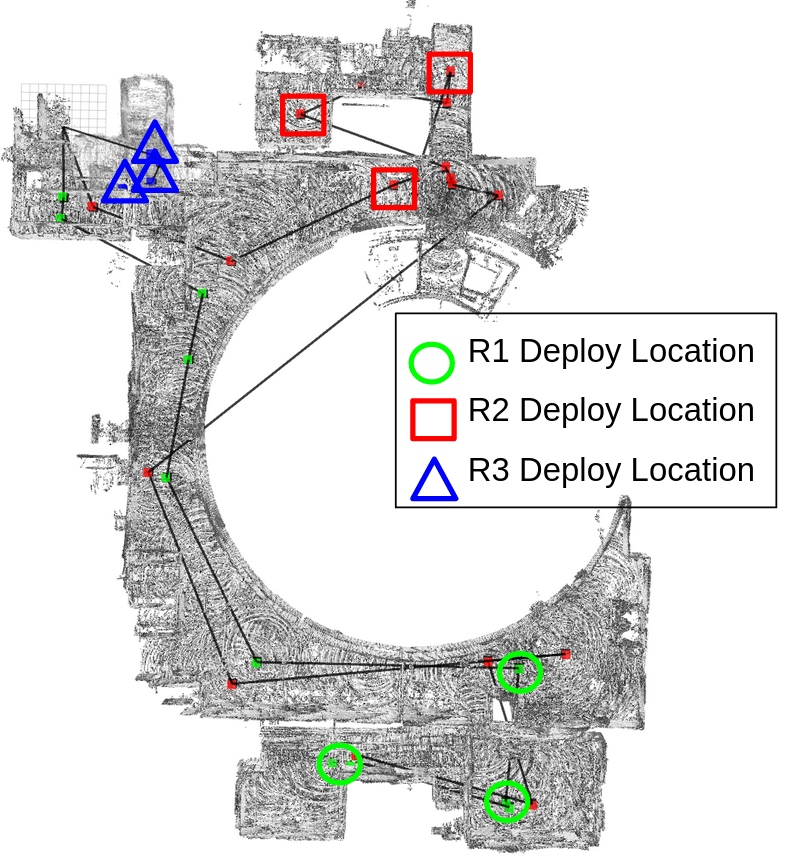}
    \caption{Deployment locations for each carrier robot in the Alpha environment from the DARPA Subterranean Urban Circuit selected by \textit{MCTS SSAP Rollout} with 10,000 iterations and $N=19$ stages. Robots 1 and 2 deploy in non-conflicting locations, avoiding the overlap conflicts. Robot 3 explores down a stairwell in the top left corner of the map and selects to deploy in those locations. The \textit{Single Robot SSAP} algorithm decided to deploy in high value but conflicting locations.}
    \vspace{-1ex}
    \label{fig:real_alpha}
\end{figure}


Our deployment algorithm was able to handle the potential deployment conflicts between the $R=3$ robots and captured more features than the competing deployment algorithms. The \textit{MCTS SSAP Rollout} algorithm captured 157 features, compared to \textit{Single Robot SSAP}: 122 features, \textit{MCTS Random Rollout}: 114 features, and \textit{Random}: 68 features. The selected deployment locations for each robot are illustrated in Fig.~\ref{fig:real_alpha}. The set of deployment locations achieve good spatial coverage across the environment while targeting the areas with high aerial robot exploration value. As this was a real-world mission, multi-robot coordination planning was employed between the carrier robots, which naturally seeks to prevent path overlap between the robots to maximize exploration efficiency. Compared to the simulated data in Sec.~\ref{sec:simulated_dat}, there is expected to be less non-uniformly distributed overlaps. However, there were several valuable locations that \textit{Single Robot SSAP} selects to deploy at but without consideration for overlap conflicts, leading to penalties.

\section{Future Work}
We proposed an algorithm for planning the deployment of multiple passenger robots from multiple carrier robots in a multi-marsupial robot system. Our algorithm reasons over the future value of decisions and potential conflicts between multiple deployments. In the future, we are interested in developing decentralized algorithms, such as by employing the SSAP rollout policy within \mbox{Dec-MCTS}~\cite{best2019decmcts}, to enable deployment planning in communication sparse environments. 
It would also be interesting to plan over the space of closed-loop policies, rather than open-loop action sequence, which may be achieved with an MCTS generalization such as POMCP~\cite{silver2011pomcp}, as noted in Sec.~\ref{sec:analysis}.
Furthermore, extensions for heterogeneous teams of passenger robots should be developed, which may be achieved through generalized assignment values~\cite{derman1972sequential}.
Also, different forms of conflicts and dependencies may exist between the deployment location rewards, which would lead to alternative reward function definitions, such as a set cover formulation. We would like to demonstrate the viability of our approach in other domains, such as in marine monitoring scenarios where multiple underwater robots are deployed from multiple surface vessels.


\balance






\bibliographystyle{IEEEtran}
\bibliography{main}

\begin{thebibliography}{10}
\providecommand{\url}[1]{#1}
\csname url@samestyle\endcsname
\providecommand{\newblock}{\relax}
\providecommand{\bibinfo}[2]{#2}
\providecommand{\BIBentrySTDinterwordspacing}{\spaceskip=0pt\relax}
\providecommand{\BIBentryALTinterwordstretchfactor}{4}
\providecommand{\BIBentryALTinterwordspacing}{\spaceskip=\fontdimen2\font plus
\BIBentryALTinterwordstretchfactor\fontdimen3\font minus
  \fontdimen4\font\relax}
\providecommand{\BIBforeignlanguage}[2]{{%
\expandafter\ifx\csname l@#1\endcsname\relax
\typeout{** WARNING: IEEEtran.bst: No hyphenation pattern has been}%
\typeout{** loaded for the language `#1'. Using the pattern for}%
\typeout{** the default language instead.}%
\else
\language=\csname l@#1\endcsname
\fi
#2}}
\providecommand{\BIBdecl}{\relax}
\BIBdecl

\bibitem{moore2016nested}
J.~Moore, K.~C. Wolfe, M.~S. Johannes, K.~D. Katyal, M.~P. Para, R.~J. Murphy,
  J.~Hatch, C.~J. Taylor, R.~J. Bamberger, and E.~Tunstel, ``Nested marsupial
  robotic system for search and sampling in increasingly constrained
  environments,'' in \emph{Proc. IEEE Int. Conf. on Systems, Man, and
  Cybernetics (SMC)}, 2016, pp. 2279--2286.

\bibitem{stankiewicz2018motion}
P.~G. Stankiewicz, S.~Jenkins, G.~E. Mullins, K.~C. Wolfe, M.~S. Johannes, and
  J.~L. Moore, ``A motion planning approach for marsupial robotic systems,'' in
  \emph{Proc. IEEE/RSJ Int. Conf. on Intelligent Robots and Systems (IROS)},
  2018, pp. 1--9.

\bibitem{zhao2008research}
J.~Zhao, G.~Liu, Y.~Liu, and Y.~Zhu, ``Research on the application of a
  marsupial robot for coal mine rescue,'' in \emph{Proc. Int. Conf. on
  Intelligent Robotics and Applications}, 2008, pp. 1127--1136.

\bibitem{das2015data}
J.~Das, F.~Py, J.~B. Harvey, J.~P. Ryan, A.~Gellene, R.~Graham, D.~A. Caron,
  K.~Rajan, and G.~S. Sukhatme, ``Data-driven robotic sampling for marine
  ecosystem monitoring,'' \emph{Int. J. Robotics Research}, vol.~34, no.~12,
  pp. 1435--1452, 2015.

\bibitem{kalaitzakis2020marsupial}
M.~Kalaitzakis, B.~Cain, N.~Vitzilaios, I.~Rekleitis, and J.~Moulton, ``A
  marsupial robotic system for surveying and inspection of freshwater
  ecosystems,'' \emph{J. Field Robotics}, vol.~38, no.~1, pp. 121--138, 2021.

\bibitem{mccammon2021ocean}
S.~McCammon, G.~Marcon~dos Santos, M.~Frantz, T.~P. Welch, G.~Best, R.~K.
  Shearman, J.~D. Nash, J.~A. Barth, J.~A. Adams, and G.~A. Hollinger, ``Ocean
  front detection and tracking using a team of heterogeneous marine vehicles,''
  \emph{J. Field Robotics}, vol. early access, pp. 854--881, 2021, {DOI}:
  10.1002/rob.22014.

\bibitem{Murphy1999}
R.~R. Murphy, M.~Ausmus, M.~Bugajska, T.~Ellis, T.~Johnson, N.~Kelley,
  J.~Kiefer, and L.~Pollock, ``{Marsupial-like mobile robot societies},''
  \emph{Proc. Int. Conf. on Autonomous Agents}, pp. 364--365, 1999.

\bibitem{lee2021optimalSSAP}
C.~Y.~H. Lee, G.~Best, and G.~A. Hollinger, ``Optimal sequential stochastic
  deployment of multiple passenger robots,'' in \emph{Proc. IEEE Int. Conf. on
  Robotics and Automation (ICRA)}, Virtual, 2021.

\bibitem{derman1972sequential}
C.~Derman, G.~J. Lieberman, and S.~M. Ross, ``A sequential stochastic
  assignment problem,'' \emph{Management Science}, vol.~18, no.~7, pp.
  349--355, 1972.

\bibitem{browne2012survey}
C.~B. Browne, E.~Powley, D.~Whitehouse, S.~M. Lucas, P.~I. Cowling,
  P.~Rohlfshagen, S.~Tavener, D.~Perez, S.~Samothrakis, and S.~Colton, ``A
  survey of {M}onte {C}arlo tree search methods,'' \emph{IEEE Trans. on
  Computational Intelligence and AI in Games}, vol.~4, no.~1, pp. 1--43, 2012.

\bibitem{kocsis2006uctB}
L.~Kocsis, C.~Szepesv{\'a}ri, and J.~Willemson, ``Improved {M}onte-{C}arlo
  search,'' University of Tartu, Tech. Rep., 2006.

\bibitem{scherer2021fr}
S.~Scherer \emph{et~al.}, ``Resilient and modular subterranean exploration with
  a team of roving and flying robots,'' \emph{[To appear] Field Robotics},
  2021.

\bibitem{wurm2013coordinating}
K.~M. Wurm, C.~Dornhege, B.~Nebel, W.~Burgard, and C.~Stachniss, ``Coordinating
  heterogeneous teams of robots using temporal symbolic planning,''
  \emph{Autonomous Robots}, vol.~34, no.~4, pp. 277--294, 2013.

\bibitem{schwager2011unifying}
M.~Schwager, D.~Rus, and J.-J. Slotine, ``Unifying geometric, probabilistic,
  and potential field approaches to multi-robot deployment,'' \emph{Int. J.
  Robotics Research}, vol.~30, no.~3, pp. 371--383, 2011.

\bibitem{schuster2020arches}
M.~J. Schuster, M.~G. M{\"u}ller, S.~G. Brunner, H.~Lehner, P.~Lehner,
  R.~Sakagami, A.~D{\"o}mel, L.~Meyer, B.~Vodermayer, R.~Giubilato
  \emph{et~al.}, ``The arches space-analogue demonstration mission: towards
  heterogeneous teams of autonomous robots for collaborative scientific
  sampling in planetary exploration,'' \emph{IEEE Robotics and Automation
  Lett.}, vol.~5, no.~4, pp. 5315--5322, 2020.

\bibitem{las2015path}
J.~C. Las~Fargeas, P.~T. Kabamba, and A.~R. Girard, ``Path planning for
  information acquisition and evasion using marsupial vehicles,'' in
  \emph{Proc. IEEE American Control Conf. (ACC)}, 2015, pp. 3734--3739.

\bibitem{mei2006deployment}
Y.~Mei, Y.-H. Lu, Y.~C. Hu, and C.~G. Lee, ``Deployment of mobile robots with
  energy and timing constraints,'' \emph{IEEE Trans. on Robotics}, vol.~22,
  no.~3, pp. 507--522, 2006.

\bibitem{hansen2018autonomous}
J.~Hansen, S.~Manjanna, A.~Q. Li, I.~Rekleitis, and G.~Dudek, ``Autonomous
  marine sampling enhanced by strategically deployed drifters in marine flow
  fields,'' in \emph{Proc. MTS/IEEE OCEANS}, 2018.

\bibitem{lee2018path}
H.~Lee and B.~Lee, ``Path planning based on probabilistic roadmap for initial
  deployment of marsupial robot team,'' \emph{Institute of Control, Robotics
  and Systems}, vol.~24, no.~1, pp. 80--89, 2018.

\bibitem{chen2019pareto}
W.~Chen and L.~Liu, ``Pareto {M}onte {C}arlo tree search for multi-objective
  informative planning.'' in \emph{Proc. Robotics: Science and Systems},
  Freiburg, Germany, 2019.

\bibitem{best2019decmcts}
G.~Best, O.~Cliff, T.~Patten, R.~R. Mettu, and R.~Fitch, ``\mbox{Dec-MCTS}:
  Decentralized planning for multi-robot active perception,'' \emph{Int. J.
  Robotics Research}, vol.~38, no. 2-3, pp. 316--337, 2019.

\bibitem{bauer2019monte}
D.~Bauer, T.~Patten, and M.~Vincze, ``{M}onte {C}arlo tree search on directed
  acyclic graphs for object pose verification,'' in \emph{Proc. Int. Conf. on
  Computer Vision Systems}, 2019, pp. 386--396.

\bibitem{corah2019distributed}
M.~Corah and N.~Michael, ``Distributed matroid-constrained submodular
  maximization for multi-robot exploration: Theory and practice,''
  \emph{Autonomous Robots}, vol.~43, no.~2, pp. 485--501, 2019.

\bibitem{kartal2015stochastic}
B.~Kartal, J.~Godoy, I.~Karamouzas, and S.~J. Guy, ``Stochastic tree search
  with useful cycles for patrolling problems,'' in \emph{Proc. IEEE Int. Conf.
  on Robotics and Automation (ICRA)}, 2015, pp. 1289--1294.

\bibitem{james2017analysis}
S.~James, G.~Konidaris, and B.~Rosman, ``An analysis of {M}onte {C}arlo tree
  search,'' in \emph{Proc. AAAI Conference on Artificial Intelligence}, 2017.

\bibitem{lauri2015mcts}
M.~Lauri, N.~Atanasov, G.~Pappas, and R.~Ritala, ``Active object recognition
  via {M}onte {C}arlo tree search,'' in \emph{Proc. IEEE ICRA Workshop on
  Beyond Geometric Constraints}, Seattle, WA, 2015.

\bibitem{silver2011pomcp}
D.~Silver and J.~Veness, ``{M}onte-{C}arlo planning in large {POMDPs},'' in
  \emph{Advances in Neural Information Processing Systems 23}, J.~D. Lafferty,
  C.~K.~I. Williams, J.~Shawe-Taylor, R.~S. Zemel, and A.~Culotta, Eds.\hskip
  1em plus 0.5em minus 0.4em\relax Curran Associates, Inc., 2010, pp.
  2164--2172.

\bibitem{allen_chung}
\BIBentryALTinterwordspacing
B.~Allen and T.~Chung, ``Unearthing the subterranean environment,'' 2021,
  {A}ccessed on 03.22.2021. [Online]. Available:
  \url{\url{https://www.subtchallenge.com/}}
\BIBentrySTDinterwordspacing

\bibitem{museth2013openvdb}
K.~Museth, J.~Lait, J.~Johanson, J.~Budsberg, R.~Henderson, M.~Alden, P.~Cucka,
  D.~Hill, and A.~Pearce, ``{OpenVDB}: An open-source data structure and
  toolkit for high-resolution volumes,'' in \emph{Proc. Int. Conf. on Computer
  Graphics and Interactive Techniques (SIGGRAPH)}, 2013, pp. 1--1.

\bibitem{yamauchi1997frontier}
B.~Yamauchi, ``A frontier-based approach for autonomous exploration,'' in
  \emph{Proc. IEEE Int. Symp. on Computational Intelligence in Robotics and
  Automation (CIRA)}, 1997, pp. 146--151.

\end{thebibliography}
\end{document}